\documentclass[10pt,twocolumn,letterpaper]{article}
\usepackage{multirow}
\usepackage{amsmath}
\usepackage{amssymb}
\usepackage{mathtools}
\usepackage{amsthm}
\usepackage{bm}
\usepackage{makecell}
\usepackage{float}
\usepackage[pagenumbers]{cvpr} 

\definecolor{cvprblue}{rgb}{0.21,0.49,0.74}
\usepackage[pagebackref,breaklinks,colorlinks,allcolors=cvprblue]{hyperref}

\def\paperID{***} 
\def\confName{CVPR}
\def\confYear{2026}

\title{Adaptive 1D Video Diffusion Autoencoder}

\author{
  Yao Teng\textsuperscript{1} 
  \quad 
  Minxuan Lin\textsuperscript{2} 
  \quad 
  Xian Liu\textsuperscript{3} 
  \quad 
  Shuai Wang\textsuperscript{4} 
  \quad 
  Xiao Yang\textsuperscript{2}
  \quad 
  Xihui Liu\textsuperscript{1}\thanks{Corresponding Author} \\
  \textsuperscript{1}The University of Hong Kong \quad
  \textsuperscript{2}ByteDance Inc. \quad
  \textsuperscript{3}CUHK \quad
  \textsuperscript{4}Nanjing University
}

\begin{document}

\twocolumn[{
\let\oldtwocolumn\twocolumn
\renewcommand\twocolumn[1][]{\oldtwocolumn} 
\maketitle
\begin{center}
    \includegraphics[width=\linewidth]{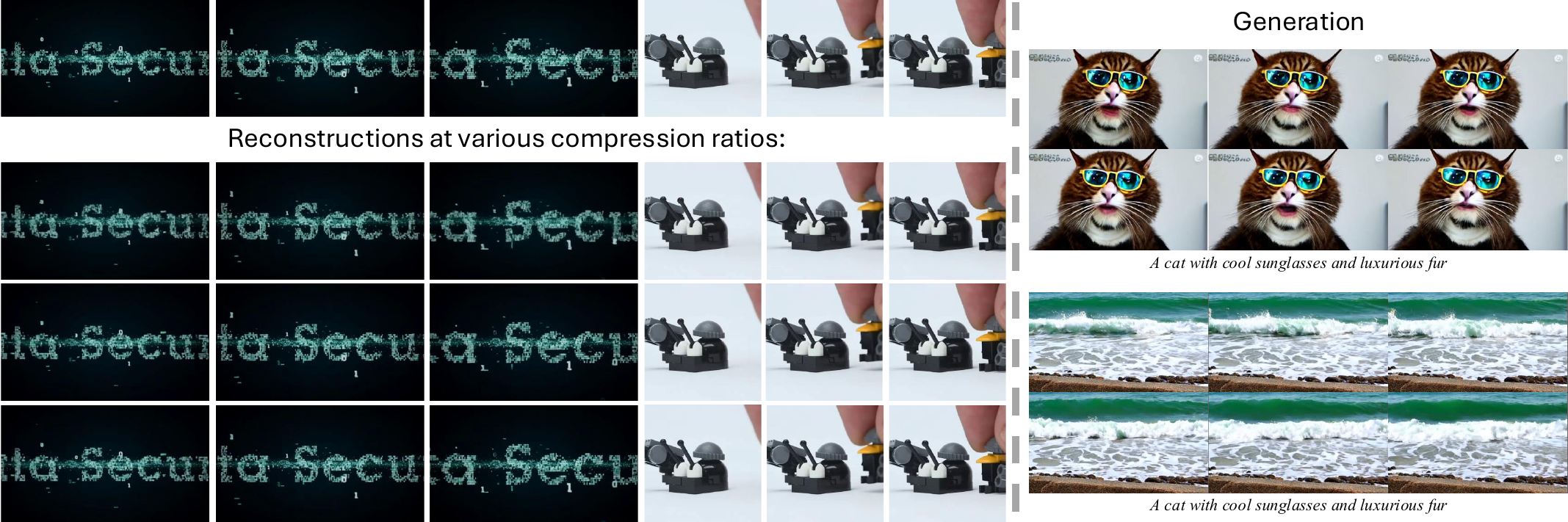}
    \captionsetup{type=figure}
    \vspace{-1.75em}
    \caption{One-Dimensional Diffusion Video Autoencoder (One-DVA). This model supports variational-length encoding, where increasing the latent length allows for the capture of richer details. Furthermore, the diffusion-based text-to-video generation can be performed on its latent space.}
    \label{fig:teaser}
    \vspace{8pt}
\end{center}
}]

\begin{abstract}
Recent video generation models largely rely on video autoencoders that compress pixel-space videos into latent representations. However, existing video autoencoders suffer from three major limitations: (1) fixed-rate compression that wastes tokens on simple videos, (2) inflexible CNN architectures that prevent variable-length latent modeling, and (3) deterministic decoders that struggle to recover appropriate details from compressed latents. To address these issues, we propose One-Dimensional Diffusion Video Autoencoder (One-DVA), a transformer-based framework for adaptive 1D encoding and diffusion-based decoding. The encoder employs query-based vision transformers to extract spatiotemporal features and produce latent representations, while a variable-length dropout mechanism dynamically adjusts the latent length. The decoder is a pixel-space diffusion transformer that reconstructs videos with the latents as input conditions. With a two-stage training strategy, One-DVA achieves performance comparable to 3D-CNN VAEs on reconstruction metrics at identical compression ratios. More importantly, it supports adaptive compression and thus can achieve higher compression ratios.
To better support downstream latent generation, we further regularize the One-DVA latent distribution for generative modeling and fine-tune its decoder to mitigate artifacts caused by the generation process.
\end{abstract}    
\section{Introduction}
\label{sec:intro}

In the field of visual generation, generative models typically rely on a pre-trained video autoencoder to facilitate the generation process. This autoencoder compresses video representations from pixel space into \textit{latents} (or tokens), enabling the generative model to generate the latents with relatively small sizes rather than handling the vast pixel data directly.
The autoencoder consists of an encoder and a decoder, and they are trained jointly. The encoder uses a neural network to compress the input videos into a latent, and the decoder reconstructs the video from this latent.

Existing video autoencoders~\citep{agarwal2025cosmos,kong2024hunyuanvideo,wan2025wan,kondratyuk2023videopoet} face several critical limitations, and we propose targeted solutions to address these challenges:
\textbf{(1) Fixed Compression Rate:} Not all videos require the same token count. For instance, a 24 fps, 5-second, 1080p video typically demands around 200,000 tokens with $16\times$ spatial and $4\times$ temporal compression. However, simple videos can be represented with far fewer tokens than the videos with complicated textures and motions. To optimize token efficiency, we propose to adopt dynamic variable-length compression, allowing adaptive latent sizes tailored to the video contents. Currently, several works~\cite{wang2024larp,tao2024learning} transform the video inputs into variable-length 1D discrete token sequences for adaptive encoding and verify this idea on class-to-video generation.
\textbf{(2) Inflexible CNN Architecture.} Convolutional neural networks (CNNs) rely heavily on human-designed priors, and their fixed-size kernels struggle to process variable-shaped inputs, limiting their ability to decode variable-length latents. In contrast, transformer architectures offer superior flexibility, processing inputs and outputs of any shape via attention mechanisms. Aligning with ``The Bitter Lesson''~\citep{sutton2019bitter}, transformers leverage large-scale data and computation to achieve greater representation capacity with minimal human priors.
\textbf{(3) Lossy Compression:} Current compression methods, whether manually determined (\eg, $16\times$ spatial and $4\times$ temporal) or dynamically estimated, aim to balance reconstruction quality and token count but struggle to achieve lossless results. When the token counts are too low, compression becomes overly lossy, requiring the decoder to infer missing details. Therefore, we deem the reconstruction as a subtask of generation and propose to use a generative decoding paradigm that allows the decoder to learn the dataset distribution and compensate omitted details through generation, minimizing reconstruction errors at the distribution level.
In summary, our research aims to design a transformer-based autoencoder with a generative decoder and variable-length compression, and successfully train this framework to match advanced autoencoders in reconstruction quality while supporting downstream generation.

In this paper, we introduce One-Dimensional Diffusion Video Autoencoder (One-DVA), a transformer-based framework~\cite{transformer} that achieves adaptive video compression and generative reconstruction within a unified design. 
The encoder leverages a Vision Transformer (ViT)~\cite{vit} that produces structural latents from spatiotemporal embeddings, while a set of 1D queries interacts with the features in transformer blocks to extract 1D latents.
A variable-length dropout mechanism is applied to the 1D latent sequence, dynamically adjusting its length to match video complexity. The decoder is implemented as a pixel-space Diffusion Transformer (DiT)~\cite{dit,wang2025pixnerd}. It treats the latents as conditional inputs and performs the diffusion process in pixel space to reconstruct the videos.

To ensure One-DVA achieves high reconstruction performance across varying compression levels, we employ a two-stage training strategy: the first stage prioritizes encoder optimization, while the second stage integrates variable-length compression and diffusion-based decoding.
With the standard compression ratio, One-DVA achieves reconstruction performance comparable to 3D-CNN VAEs. This high-fidelity reconstruction ensures that the latent space faithfully preserves the information necessary for downstream latent diffusion models (LDM). 
To further tailor the latent space for the LDM, we project the 1D latents into the space of structural latents via an alignment loss as the regularizer, facilitating joint modeling within a single LDM architecture while preserving the reconstruction performance of the autoencoder.
To ensure the visual quality of generated videos, we fine-tune the One-DVA decoder using latents generated by the LDM. These latents serve as noisy inputs that help the decoder adapt to the potential artifacts produced by the process of latent generation.

\section{Background and Related Work}
\label{sec:related}

\paragraph{Image Autoencoders}
In this paper, we classify image autoencoders based on the following attributes: 
\textbf{(1) Latent Representation Type}: continuous or discrete; 
\textbf{(2) Latent Shape}: 2D or 1D; 
\textbf{(3) Architecture}: CNN or transformer~\cite{transformer,vit}; 
\textbf{(4) Decoder Paradigm}: deterministic or generative.
In the following paragraphs, we will discuss existing autoencoders categorized by these attributes.

\noindent
\textit{\textbf{Continuous 2D CNN Autoencoder}}:
The most classic image autoencoder is the continuous 2D CNN autoencoder~\cite{dai2019diagnosing,chen2024deep,chen2025dc}. The encoder accepts an input image and outputs a low-dimensional 2D latent map through a CNN. This latent map has reduced height and width but a slightly larger channel dimension compared to the input image. The CNN decoder then takes the 2D latent map as input and reconstructs it into an image. Latent diffusion models~\cite{stable_diffusion,sdxl,dit,sit,uvit,sd3,teng2024dim,dis,zigma,wang2024flowdcn,wang2025ddt} perform the diffusion process in the latent space.

\noindent
\textit{\textbf{Discrete 2D CNN Autoencoder}}:
The discrete 2D CNN autoencoder (visual tokenizer)~\cite{vqgan,yu2023magvit,yu2023language,luo2024open-magvit2,wang2025end,zhang2025quantize} is characterized by employing quantization strategies (such as VQ~\cite{vqgan}, RQ~\cite{lee2022autoregressive}, FSQ~\cite{mentzer2023finite}, or BSQ~\cite{zhao2024image-bsq}) to convert continuous latents into discrete tokens. Generative models, such as autoregressive models~\cite{dalle,ding2021cogview,yu2022scaling-parti,tian2024var,wang2025parallelized,liu2024lumina-mgpt,sun2024llamagen,chern2024anole,Emu-3,wang2025simplear}, then learn to generate these discrete tokens to represent an image.

\noindent
\textit{\textbf{2D Transformer Autoencoder}}:
Having discussed 2D CNN autoencoders, we now turn to a new group of 2D autoencoders that are mainly built on transformer blocks. These autoencoders typically contain a ViT~\cite{vit} in their encoder. The transformer architecture enables the use of pre-trained foundational models (such as CLIP~\cite{clip,zhai2023sigmoid,tschannen2025siglip} or DINO~\cite{oquab2023dinov2,simeoni2025dinov3}) as the main component of the encoder~\cite{zheng2025vision-foundation,li2025manzanosimplescalableunified,ma2025unitok,chen2025aligningvisualfoundationencoders,tang2025unilip,zheng2025diffusion,shi2025latent}. Additionally, the scalability of the transformer-based models in other domains prompts the exploration of their potential in image reconstruction tasks~\cite{hansen2025learnings,xiong2025gigatok}.

\noindent
\textit{\textbf{1D Autoencoder}}: Images can also be represented by 1D latents or tokens, in addition to 2D ones. 1D autoencoders~\cite{yu2024image,duggal2024adaptive,xiong2025gigatok,wu2025alitok,miwa2025one,beyer2025highly,liu2025detailflow,qiu2025image,yang2025latent,chen2025masked,chen2025softvq,zha2025language,kim2025democratizing} typically adopt query-based transformer architectures~\cite{detr,deformabledetr,sparsercnn,adamixer,stageinteractor,wang2023deep-deqdet,structured_sparse_rcnn,li2023blip,ge2023planting-seed} due to their flexibility. In the encoder, the 1D learnable queries extract features from the input images by the attention mechanism~\cite{transformer}, producing 1D continuous latents (or discrete tokens). In the decoder, another learnable vector is used to reconstruct the input image. This vector is repeated to match the shape of the input image and is then fed into another transformer to retrieve image information from the latent features, thereby fulfilling the reconstruction task. 
Since the latent shape can be arbitrarily determined in this autoencoder with 1D being the simplest, we can modify the compression ratio by changing the quantity of the 1D learnable queries.
Specifically, during training, dynamic compression can be achieved through variable query counts using the tail dropout, resembling matryoshka learning~\cite{kusupati2022matryoshka}. The dropout length is randomly sampled from the uniform distribution~\cite{miwa2025one} or estimated by learnable scorers~\cite{yan2024elastictok,tao2024learning}.

\noindent
\textit{\textbf{Autoencoder with Diffusion Decoder}}: Some approaches replace the deterministic decoder with a diffusion-based decoder~\cite{gao2025dar,pan2025generative,sargent2025flow,bachmann2025flextok,wen2025principal,chen2025diffusion}. For 1D autoencoders, the most intuitive way is directly substituting the learnable vector of the decoder with random Gaussian noise, enabling conditional pixel-space diffusion generation, where latents or tokens serve as the conditions injected by attention~\cite{gao2025dar,pan2025generative,sargent2025flow,bachmann2025flextok,wen2025principal}. For 2D CNN autoencoders, a Gaussian noise branch is added, with condition injection via ControlNet~\cite{controlnet} or channel concatenation~\cite{zhao2024epsilon}.

\vspace{-1em}
\paragraph{Video Autoencoders}
Similar to image autoencoders, video autoencoders also adopt 3D CNN-based continuous~\cite{kong2024hunyuanvideo,wan2025wan,hacohen2024ltx,cheng2025leanvae,wang2025vidtwin,li2025wf,yang2024cogvideox,agarwal2025cosmos} and discrete~\cite{yu2023language,wang2024omnitokenizer,tang2024vidtok,agarwal2025cosmos} frameworks, which support diffusion-based~\cite{latte,sora,kong2024hunyuanvideo,wan2025wan,lin2024open,ma2025step,teng2025magi,zhang2025waver,deng2025magref,guo2024i2v} and autoregressive-based~\cite{kondratyuk2023videopoet,wang2024loong} video generation, respectively.
In addition, 1D autoencoders~\cite{wang2024larp,tao2024learning} and 3D transformer-based autoencoders~\cite{teng2025magi,lu2025atoken,liu2025hi} have been employed. Diffusion-based video autoencoder decoders have emerged~\cite{yang2025rethinking,zhang2025regen,liu2025hi}.
Although the overall framework designs of video and image autoencoders are similar, their latent representations differ in structure. The advanced video autoencoders~\cite{kong2024hunyuanvideo,wan2025wan,agarwal2025cosmos} typically employ a first-frame plus temporal compression strategy, which transforms an input video with a shape of $\mathcal{T} \times \mathcal{H} \times \mathcal{W}$ into a compressed latent with a shape of $\left(1 + \lceil \frac{\mathcal{T}-1} {\mathcal{P}_t} \rceil \right) \times \lceil \frac{\mathcal{H}}{\mathcal{P}_s} \rceil \times \lceil \frac{\mathcal{W}}{\mathcal{P}_s} \rceil $. This design achieves $\mathcal{P}_s \times$ compression in the spatial dimensions, $\mathcal{P}_t \times$ compression in the temporal dimension, and an additional pure spatial compression for the first frame.

\vspace{-1em}
\paragraph{Discussion}
Compared to prior works, this paper proposes to integrate the following three key features into one video autoencoder, and we train this model to achieve reconstruction performance comparable to existing autoencoders:
(1) \textit{\textbf{1D variable-length encoding}} that enables dynamic compression ratios.
(2) A query-based \textit{\textbf{transformer}} architecture that allows for flexible video information extraction, forming the foundation for the variable-length encoding.
(3) A \textit{\textbf{diffusion decoder}} that improves reconstruction quality.

\section{Method}

This paper proposes the One-Dimensional Diffusion Video Autoencoder (One-DVA), a transformer-based framework that supports variable-length 1D encoding and pixel-space diffusion decoding. As illustrated in~\cref{fig:tokenizer}, the autoencoder consists of an encoder, a latent-dropout module, and a diffusion-based decoder. The encoder compresses a video into two complementary representations: a structural latent obtained from the ViT backbone, and a 1D latent sequence extracted via a query mechanism. The decoder reconstructs the original frames through conditional pixel-space video diffusion. The condition is formed by the structural latent and the 1D latents, and the 1D latents can be truncated via dropout to achieve variable length.

\subsection{Query-based Vision Transformer Encoder}
\label{sec:encoder}
\begin{figure}[t]
\centering
\includegraphics[width=1\linewidth]{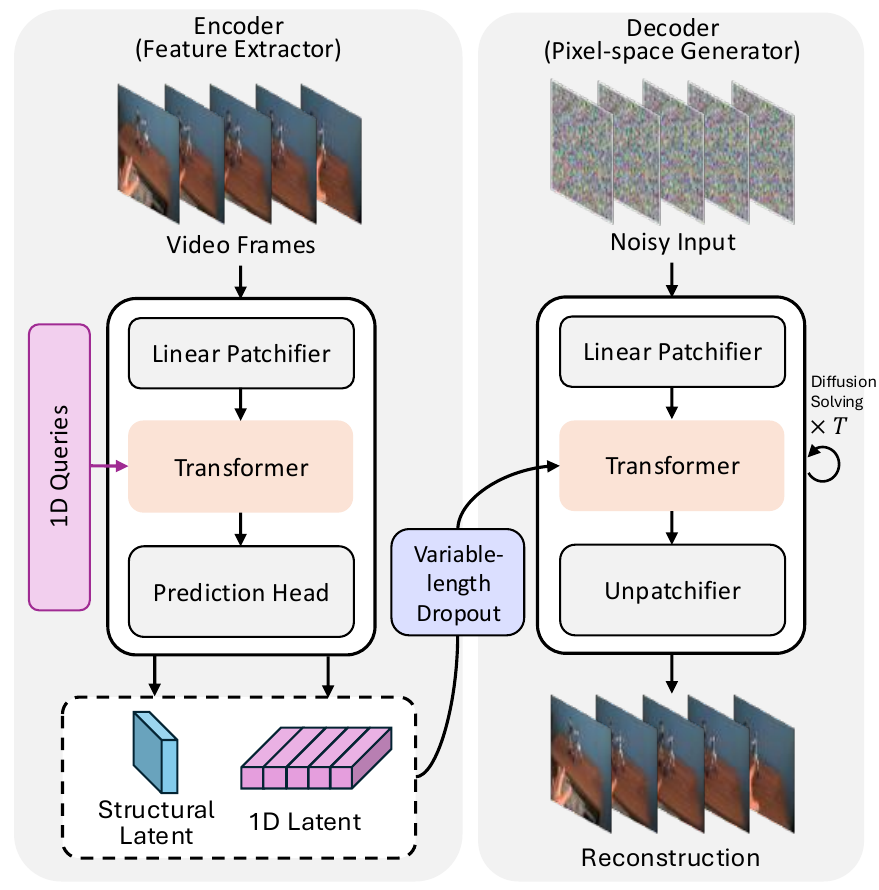}
\vspace{-2em}
\caption{
\textbf{Overview:} our One-DVA consists of an encoder, a diffusion decoder and a latent dropout module. The encoder utilizes a vision transformer with 1D queries to extract input video features and outputs low-dimensional latents.
The latent dropout module dynamically adjusts the length of 1D latents during training.
The diffusion decoder is a diffusion transformer generating videos in pixel space with the latents as the input condition.
} 
\label{fig:tokenizer}
\end{figure}

The transformer architecture is particularly well-suited for encoding videos into variable-length latents, as it processes all inputs as token sequences and applies self-attention in a unified manner, accommodating arbitrary shapes with ease.
As illustrated in \cref{fig:tokenizer}, the input video frames are first processed by a linear patchifier, which projects the RGB pixels into high-dimensional spatiotemporal embeddings. These embeddings are then flattened into a sequence and concatenated with learnable 1D queries before being fed into a stack of transformer blocks. For these sequential queries, we apply learnable positional encodings on them, following~\cite{yu2024image}. For the spatiotemporal embeddings, absolute positional encodings are used. To facilitate multi-resolution training, we further concatenate special tokens representing the height, width, temporal length, spatial size, and spatial aspect ratio of the inputs to the sequence, in line with~\cite{pixartalpha}.
Through the self-attention mechanisms in the transformer blocks, the spatiotemporal embeddings are processed into features and the queries selectively extract the essential spatiotemporal visual content required for reconstruction. Subsequent to the transformer blocks, both the processed spatiotemporal features and the 1D query features are passed through a channel compression layer to reduce their channel dimensions.
Since the total number of query tokens and spatiotemporal feature vectors exceeds the latent size in standard video encoding (detailed in \cref{sec:related}), we select a subset of them to form the final latent.
Specifically, given a video input of shape $\mathcal{T} \times \mathcal{H} \times \mathcal{W}$, we obtain the 1D latents by selecting the first $\left( \lceil \frac{\mathcal{T}-1}{\mathcal{P}_t}  \rceil \times \lceil \frac{\mathcal{H}}{\mathcal{P}_s} \rceil \times \lceil \frac{\mathcal{W}}{\mathcal{P}_s} \rceil \right)$ queries.
Subsequently, we derive a structural latent of size $(1 \times \lceil \frac{\mathcal{H}}{\mathcal{P}_s} \rceil \times \lceil \frac{\mathcal{W}}{\mathcal{P}_s} \rceil)$ by sampling the channel-compressed spatiotemporal features from the ViT and performing spatial downsampling. 
In total, the input video is thus represented by a hybrid latent of shape $\left(1 + \lceil  \frac{\mathcal{T}-1}{\mathcal{P}_t} \rceil \right) \times \lceil \frac{\mathcal{H}}{\mathcal{P}_s} \rceil \times \lceil \frac{\mathcal{W}}{\mathcal{P}_s} \rceil$, consistent with advanced video autoencoders~\cite{kong2024hunyuanvideo,wan2025wan,agarwal2025cosmos}.

\subsection{Variable-length Encoding}
\label{sec:varlen}

As illustrated in \cref{fig:tokenizer}, the variable-length dropout module dynamically adjusts the 1D latent length. This is achieved via a ``matryoshka'' training strategy~\cite{kusupati2022matryoshka}. During training, the module applies random dropout to the 1D latents starting from the tail toward the head to vary their length, with the dropout ratio sampled from a distribution governed by a motion score computed from pixel differences (see the appendix for details). In the decoder, the dropped tokens are replaced with padding tokens. Furthermore, we configure $10\%$ of conditions to use full latents and another $10\%$ to employ only structural latents. Through this variable-length dropout mechanism, the generative models are able to learn to generate latents of different sizes.

\subsection{Diffusion Decoding}
While compression ratios can be estimated or manually tuned, lossy compression inherently risks the reconstruction error. To enhance reconstruction quality, we consider the decoder to possess generative capabilities. Specifically, we treat the decoding process as a conditional generation task, using variable-length token sequences as conditions within a diffusion-based generative framework.
As illustrated in \cref{fig:tokenizer}, the decoder takes two inputs: a condition (i.e., variable-length token sequences processed by the sampler) and a noisy input (either random noise or the encoder input perturbed by random noise). During diffusion training, the noisy input is obtained by perturbing the ground-truth video with noise:
\begin{equation}
\bm{x}_t = (1 - t) \cdot \bm{x}_0 + t \cdot \bm{x}_1, \quad \bm{x}_1 \sim \mathcal{N}(\bm{0}, \bm{I}),
\end{equation}
where $\bm{x}_0$ represents the ground-truth video clip, $t$ is a sampled timestep ranging from $0$ to $1$ (timestep sampling details in~\cref{sec:train}), and $\bm{x}_1$ is random Gaussian noise.
During inference, diffusion sampling progressively refines a random noise input into a clean video:
\begin{equation}
\bm{x}_t = \bm{x}_s + \mathcal{D}_{\theta}(\bm{x}_s, s, \bm{z}) \cdot (t - s), \quad \bm{x}_1 \sim \mathcal{N}(\bm{0}, \bm{I}),
\end{equation}
where $\mathcal{D}_{\theta}$ denotes the decoder output, i.e., the velocity prediction, $\bm{c}$ represents the latents (more details of $\bm{z}$ are in~\cref{sec:train}), and $t$ and $s$ are consecutive timesteps with $t < s$ and $s$ starting from $1$.

\vspace{-1em}
\paragraph{Decoder Architecture}
Our decoder is a pixel diffusion transformer~\cite{wang2025pixnerd}.
The latent input and noisy input are first transformed into high-dimensional features via linear layers, then concatenated and fed into the transformer blocks. For the output of the transformer, the features corresponding to the noisy input positions are separated from the output sequence.
Inspired by~\cite{uvit,teng2025magi,wang2025pixnerd}, our unpatchifier consists of a long skip connection, a linear projection, a pixel-shuffle operation, and a final convolutional layer.

\subsection{Autoencoder Training}
\label{sec:train}

\paragraph{Loss Functions}
As the decoder employs a diffusion-based paradigm, we use a diffusion loss (implemented as flow-matching loss) to train our autoencoder rather than the commonly used reconstruction loss. Specifically, the diffusion loss is as follows:
\begin{equation}
\mathcal{L}_{\text{diff}} = \mathbb{E}_{t, \bm{x}_1, \bm{x}_0} \left[ \left\| \mathcal{D}_{\theta} \left( \bm{x}_t, t, \bm{z} \right) - \left( \bm{x}_1 - \bm{x}_0 \right) \right\|_2^2 \right].
\label{eq:diff}
\end{equation}
The training optimizes a composite loss function:
\begin{equation}
\mathcal{L} = \lambda_1 \mathcal{L}_{\text{diff}} + \lambda_2 \mathcal{L}_{\text{perceptual}} + \lambda_3 \mathcal{L}_{\text{kl}} + \lambda_4 \mathcal{L}_{\text{repa}},
\end{equation}
where $\mathcal{L}_{\text{diff}}$ is the diffusion loss defined in~\cref{eq:diff}, $\mathcal{L}_{\text{perceptual}}$ is the perceptual loss~\cite{johnson2016perceptual} between VGG features of real and reconstructed frames, $\mathcal{L}_{\text{kl}}$ is the KL loss~\cite{kingma2013auto} that regularizes the latents to satisfy the standard Gaussian distribution, $\mathcal{L}_{\text{repa}}$ is the REPA loss~\cite{yu2024representation} performed on the features of noisy inputs in the decoder~\cite{xiong2025gigatok,gao2025dar}, and $\lambda_1, \lambda_2, \lambda_3, \lambda_4$ are weighting coefficients.

\vspace{-1em}
\paragraph{Training Recipe} We empirically observe that a multi-stage training procedure is more effective in training our autoencoder well than the end-to-end training:

\noindent
\textbf{Stage 1: Deterministic Pretraining.}
This stage focuses on training the encoder to extract features critical for reconstruction. To avoid information leakage that would simplify the reconstruction task, we input pure random noise (\ie, $t \equiv 1$) into the decoder. This forces the encoder to capture all essential information required for reconstruction.
Additionally, we disable variable-length dropout to establish the upper bound of the reconstruction ability of One-DVA. Thus, the latent inputs for the decoder are set as $\bm{c} = \mathcal{E}_{\phi}(\bm{x}_0)$ where $\mathcal{E}_{\phi}$ denotes the encoder.
In this configuration, our autoencoder behaves more like an end-to-end model than a diffusion model, as it does not support the multi-step denoising in principle.

\noindent
\textbf{Stage 2: Stochastic Post-Training.}
We unleash the diffusion timestep sampling and variable-length dropout in this stage. Following~\cite{sargent2025flow}, we adopt a thick-tailed logit-normal sampling for diffusion timesteps and thus we sample a noise level as full noise at $10\%$ of the time. 
Then, we introduce variable-length compression by applying dropout to the latents: $\bm{z} = \text{Dropout} \left( \mathcal{E}_{\phi}(\bm{x}_0), l \right)$, where $l$ is the dropout ratio defined in~\cref{sec:varlen}.

\noindent
With this two-stage training strategy, we can train an autoencoder with high reconstruction fidelity. However, it does not inherently guarantee a latent space or latent-to-pixel decoder optimized for downstream diffusion-based video generation. In the following section, we describe another post-training stage to adapt the autoencoder for the generation tasks.

\subsection{Adapting Autoencoder for Video Generation}
\label{sec:dit}

We train latent diffusion models (LDM) on the latent space of One-DVA. Formally, the latent space exhibits a clear separation between the structural latents and 1D latents, derived from the spatiotemporal patches and learnable queries with dropout, respectively. While diffusion modeling on structural latents from ViT has been validated~\cite{teng2025magi}, that of the variational 1D latents remains under-explored. 
To achieve high-quality video synthesis on such latent space, we propose latent space alignment for joint modeling and fine-tune the decoder using LDM-sampled latents to suppress generation artifacts.

\paragraph{Latent Space Alignment}

The spatiotemporal patches in ViT naturally exhibit locality and spatial structural priors~\cite{oquab2023dinov2,simeoni2025dinov3}, where adjacent latent vectors are identically distributed and show high local similarity. These attributes are essential for efficient diffusion learning~\cite{singh2025matters,liu2025delving}. In contrast, learnable queries lack predefined positional information. While highly flexible, they offer no inherent structural guaranties. To address this issue, we inject structural priors into the 1D latents through a self-alignment mechanism. Specifically, for each video, we align each 1D latent vector with its best-matching counterpart in the structural latent vector by minimizing their top-1 cosine distance. Additionally, we enforce internal continuity by maximizing the self-similarity between each 1D latent vector and its nearest neighbor. 
Integrating this regularization, we further fine-tune One-DVA for additional iterations. Empirically, this regularization maintains reconstruction fidelity without degradation, given a proper loss weight.

\vspace{-1em}
\paragraph{Decoder Fine-tuning} The sampling process of generative models inevitably introduces prediction errors~\cite{zhang2025waver,wang2025pixnerd}. In our framework, this manifests as a distributional drift~\cite{bengio2015scheduled} between encoded and predicted latents, leading to noticeable patch-like artifacts in the pixel space. This drift could lack closed form. To bridge this training-inference gap, we directly fine-tune the decoder using predicted latents, following the intuition of~\cite{qiu2025image}. Specifically, we optimize the decoder to reconstruct the original ground-truth videos by taking the latents sampled from our LDM as input, rather than the encoded ones. We freeze the encoder during this process to keep the latent space stationary. Empirically, this strategy effectively eliminates generation artifacts within a few thousand iterations.

\begin{table*}
    \centering
    \begin{tabular}{l|ccc|cccc}
    \toprule
    Autoencoders & \makecell{Compr. Ratio\\$(\mathcal{P}_t \times \mathcal{P}_s \times \mathcal{P}_s)$} & \makecell{Channel\\Dim} & \makecell{Auxiliary\\ Compr. Ratio} & rFVD ($\downarrow$) & PSNR ($\uparrow$) & SSIM ($\uparrow$) & LPIPS ($\downarrow$) \\
    \midrule
    CogVideoX~\cite{yang2024cogvideox} & $4 \times 8 \times 8 $ & $ 16$ & $ 8 \times 8 $  & 68.17 & 34.97 &  0.94 & 0.033 \\
    HunyuanVideo~\cite{kong2024hunyuanvideo} & $4 \times 8 \times 8$ & $ 16$ & $8 \times 8$  & \textbf{51.47}  & 35.54 & 0.94 & \textbf{0.023} \\
    Wanx2.1~\cite{wan2025wan} & $4 \times 8 \times 8$ & $ 16$ & $8 \times 8$ & 62.25 & 34.95 & 0.94 & 0.024 \\
    Wanx2.2~\cite{wan2025wan} & $4 \times 16 \times 16$ & $ 48$ & $16 \times 16$ & 60.18 & 35.23 &  0.94 & 0.023 \\
    Magi1~\cite{teng2025magi} & $4 \times 8 \times 8$ & $ 16$ & / & 70.07 & 36.25 & \textbf{0.95} & 0.035 \\
    \midrule
    \textbf{Ours} & $ (4  \times  16  \times  16)$ & $ 64$ & $16 \times 16$ & \underline{\textbf{56.96}} & \textbf{36.48}  & \textbf{0.95} & 0.025 \\ 
    \textbf{Ours} (\textit{Avg} $55.8\%$ \textit{1D}) & $ (\frac{4 \times 16 \times 16 }{ 55.8\%} ) $  & $ 64$ & $16 \times 16$ & 70.28  & 35.42 & 0.94 & 0.029 \\ 
    \textbf{Ours} (\textit{Con} $55.8\%$ \textit{1D}) & $ (\frac{4 \times 16 \times 16 }{ 55.8\%} ) $  & $ 64$ & $16 \times 16$ & 72.42 & 35.40 &  0.94 & 0.029 \\ 
    \textbf{Ours} ($0\%$ \textit{1D}) & / & $ 64$ & $16 \times 16$ & 149.97 & 32.80 & 0.91  & 0.057 \\
    \bottomrule
    \end{tabular}
    \vspace{-.75em}
    \caption{
    Comparison of video reconstruction quality across different autoencoders. \texttt{Compr.} denotes \texttt{Compression}.
    \textbf{Bold} values indicate the best performance, while \underline{\textbf{bold-underlined}} values represent the second best. “/” denotes cases where the item is not applicable.
    \textit{Con} $X\%$ \textit{1D} means using the first $X\%$ of tokens per video. \textit{Avg} $X\%$ \textit{1D} means using a global average of $X\%$ tokens, with per-video selection based on the score defined in \cref{sec:varlen}.
    By default, $100\%$ 1D latents are used.
    }
    \vspace{-.5em}
    \label{tab:sota}
\end{table*}

\begin{table}[t]
    \centering
    \setlength{\tabcolsep}{3pt}
    \begin{tabular}{lr|cc}
    \toprule
    Method & Iters & rFVD ($\downarrow$) & PSNR ($\uparrow$) \\
    \midrule
    Pretrained & 415K & 67.56 & 36.02 \\
     + Further Training & + 85K & 67.36 & 35.85 \\
    + Diffusion Post-training & + 85K & \textbf{65.19} & \textbf{36.26} \\
    \bottomrule
    \end{tabular}
    \vspace{-.75em}
    \caption{Study on the post-training with diffusion scheduler.}
    \label{tab:abladiff}
\end{table}

\begin{table}[t]
    \centering
    \begin{tabular}{lr|cc}
    \toprule
    Method & Iters & rFVD ($\downarrow$) & PSNR ($\uparrow$) \\
    \midrule
    Stage-1 & 217K & \textbf{115.64} & \textbf{34.20} \\
    End-to-end & 217K & 230.06 & 31.03 \\
    \bottomrule
    \end{tabular}
    \vspace{-.75em}
    \caption{Study on the effectiveness of stage-wise training.}
    \label{tab:abla_stage}
\end{table}

\section{Experiments}

\subsection{Implementation Details}

Our autoencoder is trained to reconstruct videos of three typical resolutions: $17 \times 456 \times 256$, $17 \times 256 \times 456$, and $17 \times 256 \times 256$, with fps set as $24$. The FPS is set as $24$ for every video sample.
Since 1D latents lack spatial structure, which hinders the patchifying in latent diffusion models, we absorb the commonly-used $2 \times 2$ patchifying directly into the compression of our autoencoder, yielding a spatiotemporal compression rate of $4 \times 16 \times 16$. Then, we set the channel dimension of the latents as $64$.
Our training is conducted on large-scale internal data. We first trained the model for 415K iterations with a batch size of $48$, which took approximately $7$ days on $48$ 80G GPUs. We then continued the training with variational 1D latent length and the diffusion scheduler for approximately 800K iterations.
The model size of our autoencoder is 1.0B and we use FSDP~\cite{zhao2023pytorch} for training.
For text-to-video generation, each DiT has 1.3B parameters with the condition injection using cross-attention like~\cite{wan2025wan}.

\vspace{-1em}
\paragraph{Evaluation}
We evaluate the autoencoders on a random set with $1000$ video clips from the dataset proposed in~\cite{bain2021frozen} (spatiotemporal resolution $17 \times 256 \times 256$). We use the metrics and evaluation setup identical to Open-sora Plan~\cite{lin2024open}. The metrics include PSNR, the reconstruction FVD~\cite{unterthiner2018towards-fvd}, SSIM~\cite{wang2004image-ssim}, and LPIPS~\cite{zhang2018unreasonable-lpips}. To quantitatively evaluate the visual quality of the generated videos in the class-to-video task, we use the same evaluation code as in~\cite{latte}.

\begin{figure}[t]
    \centering
    \begin{subfigure}{0.45\textwidth}
        \centering
        \includegraphics[width=0.85\textwidth]{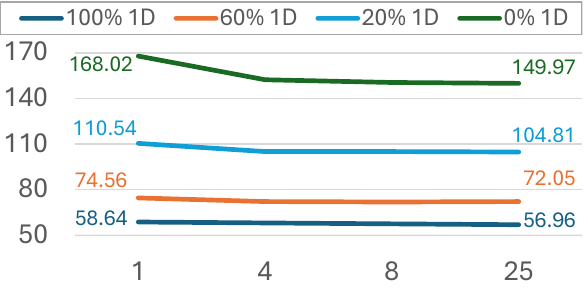}
        \vspace{-.5em}
        \caption{rFVD}
        \label{fig:1dfvd}
    \end{subfigure}
    
    \begin{subfigure}{0.45\textwidth}
        \centering
        \includegraphics[width=0.85\textwidth]{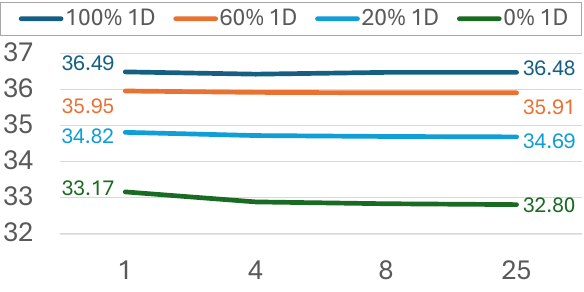}
        \vspace{-.5em}
        \caption{PSNR}
        \label{fig:1dpsnr}
    \end{subfigure}
    \vspace{-.75em}
    \caption{Reconstruction quality across different diffusion sampling steps ($1$, $4$, $8$, and $25$) and varying 1D latent lengths.}
    \vspace{-.5em}
    \label{fig:abladiffsample}
\end{figure}

\subsection{Comparison to State-of-the-art Methods}

In this section, we compare our One-DVA to the advanced autoencoders. 
Note that our autoencoder is trained on multi-resolution videos, where the total number of queries is set according to the maximal resolution to ensure compatibility. During inference on $17 \times 256 \times 256$ videos, we truncate the number of queries to achieve a \textit{standard} compression ratio, resulting in a latent length of $(4+1) \times 16 \times 16 $ with channel dimension as $64$. 
In \cref{tab:sota}, we compare our autoencoder with recent state-of-the-art video autoencoders on the task of video reconstruction. 
Firstly, our method with \textit{standard} compression ratio achieves the best overall performance in terms of PSNR and SSIM. It also attains the second-lowest rFVD score.
In the subsequent row of this table, we utilize the scoring mechanism detailed in \cref{sec:varlen} to determine the 1D latent length for each video reconstruction. In comparison, we also evaluate a baseline that applies a constant latent length across all videos with the identical usage of tokens. The results demonstrate that the reconstruction using the estimation outperforms the fixed-length approach, proving the effectiveness of our scoring strategy.
In the bottom row, in the \textit{structural-only} setting, where videos are reconstructed using solely structural latents without 1D latents, reconstruction remains feasible, albeit at the cost of reduced visual quality.

\begin{figure*}
    \centering
    \includegraphics[width=0.99\linewidth]{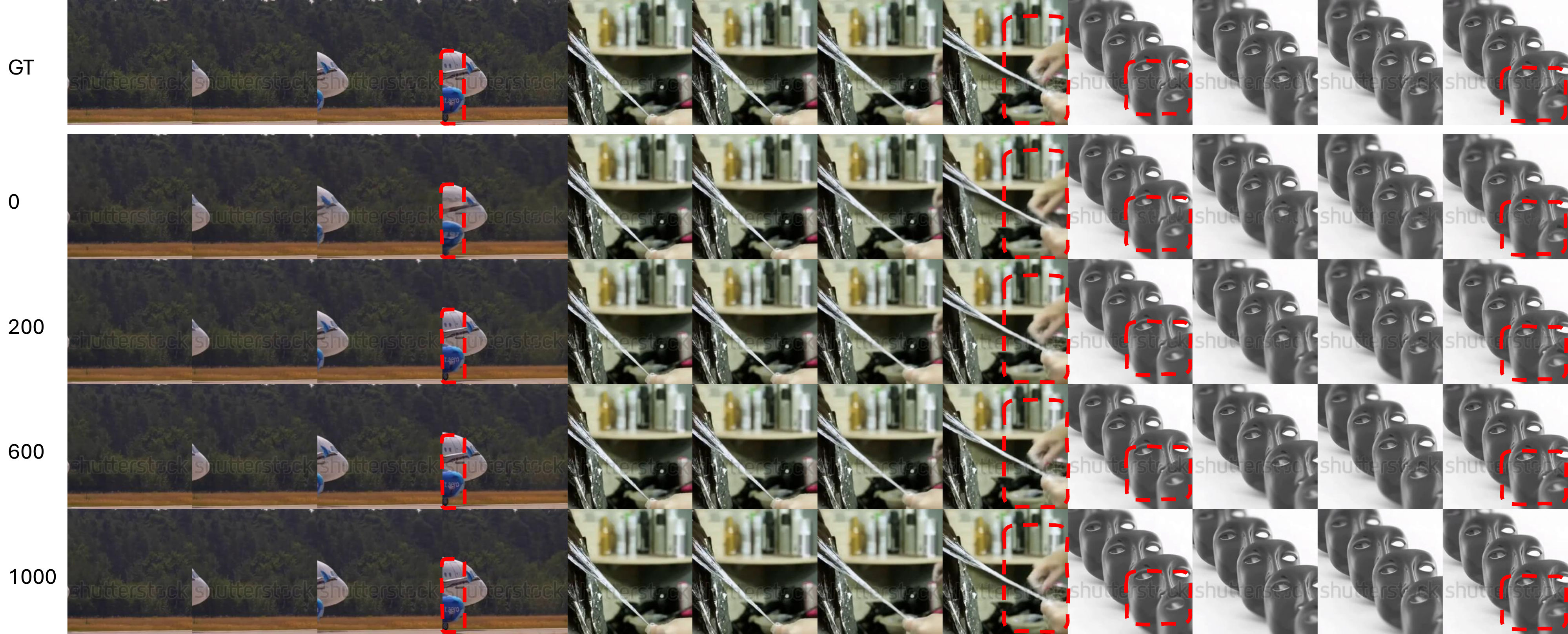}
    \vspace{-.75em}
    \caption{
    Reconstructed videos with various 1D latent lengths. The first row shows the ground-truth (GT) videos, while the subsequent rows depict reconstructions with 1D latent lengths of $0$, $200$, $600$, and $1000$, respectively. The red dashed boxes highlight regions where reconstruction quality varies noticeably across different 1D latent lengths. We sample frames at a 5-frame interval.
    }
    \vspace{-1em}
    \label{fig:recon}
\end{figure*}

\begin{figure}[t]
    \centering
    \includegraphics[width=0.99\linewidth]{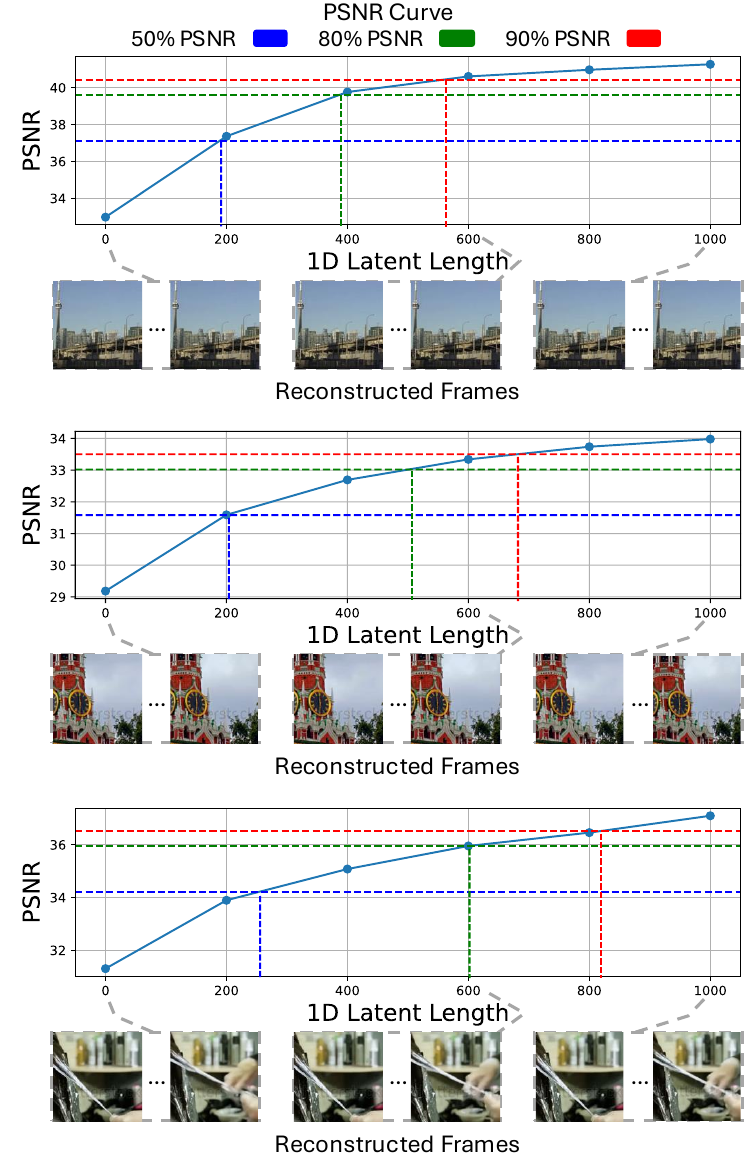}
    \vspace{-.5em}
    \caption{
    Quantitative reconstruction metrics using variable-length 1D latents. Videos with greater motion exhibit a steeper PSNR decline as the 1D latent length decreases.
    }
    \vspace{-3em}
    \label{fig:var1d}
\end{figure}

\begin{figure*}
    \centering
    \includegraphics[width=0.99\linewidth]{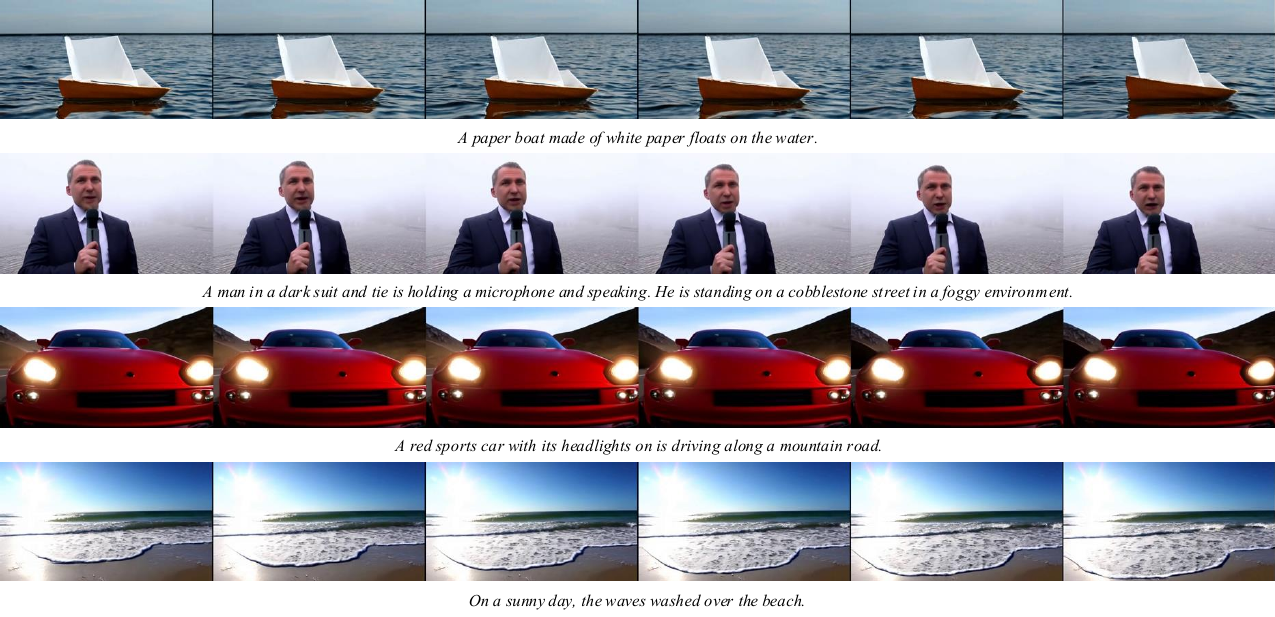}
    \vspace{-.5em}
    \caption{Text-to-video results of our latent diffusion model trained on the latent space of our autoencoder.}
    \vspace{-.5em}
    \label{fig:ldm}
\end{figure*}

\subsection{Analysis on Reconstruction}
\label{ana:recon}

\paragraph{Reconstruction with Variable-length 1D Latents}
In addition to the results in \cref{tab:sota}, which demonstrate that our autoencoder can reconstruct videos at different compression ratios, we conduct a detailed case analysis of the impact of 1D latent length on reconstruction quality. As illustrated in \cref{fig:var1d}, videos containing larger motions exhibit a steeper decline in PSNR as the 1D latent length decreases. For example, the chart shows that achieving 90\% PSNR requires a longer 1D latent length for videos with more motion. 
Moreover, we present qualitative results in \cref{fig:recon}. We observe that longer 1D latents enable more accurate reconstruction of fine details, such as scene text, whereas the video regions that contain motions appear blurry when they are reconstructed without 1D latents.

\vspace{-1em}
\paragraph{Study on Training Strategy}
As shown in~\cref{tab:abla_stage}, with a sufficient training duration, our two-stage training pipeline outperforms the end-to-end approach in reconstruction. These results show the advantages of the pretraining-then-post-training paradigm for our autoencoder: in end-to-end training, the information of input videos is leaked to the decoder, simplifying the reconstruction task and hindering the encoder from learning effective information. In contrast, our deterministic pretraining first compels the encoder to learn to capture features essential to reconstruction, and then the standard diffusion training is performed.

\vspace{-1em}
\paragraph{Effectiveness of Diffusion Scheduling}
We verify that diffusion-based training and sampling yield improved reconstruction quality. As shown in~\cref{tab:abladiff}, training without stochastic timesteps (resulting in one-step sampling directly from noise to video) produces slight changes in reconstruction performance.
In contrast, by employing stochastic timesteps and multi-step diffusion sampling with the same number of iterations, we observe a performance boost within a sufficient number of training iterations.
Furthermore, as shown in~\cref{fig:abladiffsample}, we conduct ablation studies on the number of diffusion sampling steps. We observe that increasing the number of steps yields great benefits to rFVD with insufficient condition (short 1D latents). When the condition is strong (\ie, using full 1D latents), the number of sampling steps has less impact. Moreover, we observe that rFVD improvements occur at the expense of PSNR. We hypothesize that the diffusion process prioritizes capturing the dataset distribution over per-sample reconstruction fidelity, thereby influencing the PSNR.

\subsection{Analysis on Generation}

To assess the generative capabilities of the latents in our autoencoder, we train latent diffusion models for video generation and evaluate them on two tasks: class-conditional generation and text-to-video generation.

For qualitative results, we present the results of text-to-video generation as well as the corresponding prompts in~\cref{fig:ldm}.
For quantitative evaluation, we conduct class-conditional video generation at a $17\times 256 \times 256$ spatiotemporal resolution following the benchmark in~\cite{liu2025hi}. As reported in~\cref{tab:c2v}, our full framework utilizing the One-DVA latent space achieves a gFVD of 210.9, which matches the performance of methods such as Hi-VAE~\cite{liu2025hi}. 
The One-DVA decoder fine-tuning process specifically contributes to this result. Notably, the decoder fine-tuned on the text-to-video dataset using predicted latents from the corresponding diffusion model remains effective when applied to the class-to-video task. This suggests that prediction errors are similar across different generative tasks, allowing the error-correction capability to be transferred.
In contrast, omitting this fine-tuning step leads to a noticeable degradation in gFVD. Also, the class-to-video LDM with structural latents alone (\ie, $0\%$ 1D latents) yields reasonable generation results, as these latents encode the low-frequency components of the videos. However, due to the lack of sufficient high-frequency information, these latents impose an upper bound on generation quality, consistent with the limited reconstruction performance reported in~\cref{tab:sota}.

\begin{table}[t]
    \centering
    \begin{tabular}{l|r}
    \toprule
    Methods & gFVD ($\downarrow$) \\
    \midrule
     VideoGPT~\cite{yan2021videogpt} & 2880.6 \\
     StyleGAN-V~\cite{skorokhodov2022stylegan} & 1431.0 \\
     LVDM~\cite{he2022latent} & 372.0 \\
     Latte~\cite{latte} & 478.0 \\
     iVideoGPT~\cite{wu2024ivideogpt} & 254.8 \\
     Hi-VAE + DiT~\cite{liu2025hi} & \textbf{210.9} \\
    \midrule
    \textbf{Ours} ($0\%$ \textit{1D}) & 325.8 \\ 
    \textbf{Ours} (\textit{w/o dec ft}) & 274.2 \\
    \textbf{Ours} & \textbf{210.9} \\ 
    \bottomrule
    \end{tabular}
    \vspace{-.75em}
    \caption{The quantitative results for class-to-video generation.}
    \label{tab:c2v}
\end{table}

\section{Conclusion}
In this work, we introduce One-Dimensional Diffusion Video Autoencoder (One-DVA), a transformer-based framework that unifies adaptive 1D video tokenization and diffusion-based generative decoding. By combining query-based encoding with variable-length dropout, One-DVA supports dynamic video compression. The pixel-space diffusion decoder further enhances reconstruction with the latents as conditions. Extensive experiments validate that One-DVA is comparable to advanced 3D CNN VAEs in reconstruction. Moreover, One-DVA supports downstream latent diffusion models for video generation.

{
    \small
    \bibliographystyle{ieeenat_fullname}
    \bibliography{main}
}

\clearpage
\setcounter{page}{1}
\maketitlesupplementary

\appendix

\section{Autoencoder Details}

\paragraph{Architecture Details}
Both the encoder and decoder utilize a transformer architecture with a hidden dimension of 1152, 24 blocks, and 16 attention heads.
Following~\cite{teng2025magi}, the spatial patch size is set to $8$.
Following~\cite{assran2025v}, the temporal patch size is set to $2$ for the decoder, while we set temporal patch size as $4$ to the encoder for better efficiency.
For input video sizes not divisible by the patch sizes, we apply zero padding along the spatial axes and replicate padding along the temporal axis.
Since our autoencoder is trained to reconstruct videos of three typical resolutions ($17 \times 456 \times 256$, $17 \times 256 \times 456$, and $17 \times 256 \times 256$), we set the maximum number of queries to $1938$, corresponding to a compression ratio of $4 \times 16 \times 16$.

\paragraph{Heuristic Motion-aware Token Length Estimation}
To train our autoencoder to handle variable-length 1D latents, we employ a heuristic motion estimator to compute a motion score for each video clip, which directly determines the length of the 1D latents. We compute the raw motion score \(s\) as follows: First, video frames are converted to grayscale. We then calculate absolute pixel differences between consecutive frames. Finally, the pixel differences are averaged over all spatiotemporal dimensions to obtain the non-negative scalar score. During training, exponential moving averages of the mean \(\mu\) and standard deviation \(\sigma\) of this value are maintained online, and the raw score is normalized simply as \(
\hat{s} = \frac{s}{\mu + 3\sigma }
\) to obtain a motion score in \([0, 1]\).
The normalized motion score $\hat{s}$ determines the expected fraction of the maximum 1D latent length. To introduce stochasticity while preserving the central tendency, we sample a multiplicative factor, similar to the logit-normal sampling in~\cite{sd3}:
$\eta = 2 \cdot \operatorname{sigmoid}(z), ~ z \sim \mathcal{N}(0,1)$ but the center value is $1$. Thus, the final number of temporal tokens is computed as $\operatorname{round}\bigl(\hat{s} \cdot N_{\max} \cdot \eta \bigr)$, where $N_{\max}$ is the predefined maximum token count.

\paragraph{Training Details}
We use AdamW with \((\beta_1=0.9, \beta_2= 0.999)\) for optimization, and the weight decay is set to $10^{-4}$.
\textit{Stage 1}:
The loss weights for autoencoder training are set as $\lambda_1=10$, $\lambda_2=0.1$, $\lambda_3=1 \times 10^{-4}$, and $\lambda_4=0.1$, where $\lambda_1$ is large because we observe that the $\ell_2$-norm causes a small loss value, and we increase the loss weight for balance. We set $\lambda_3=1 \times 10^{-4}$ because the larger weight for KL loss causes the overall loss spike until the training is stable. 
For REPA loss~\cite{yu2024representation}, we use an image foundational model, SigLIP~\cite{tschannen2025siglip}, for providing supervision, because this model shows reconstruction ability proven in~\cite{chen2025blip3}. As there is temporal patchifying in our model, we interpolate the features across the temporal dimension for the supervision of REPA.
The learning rate for our first stage training is set to \(5 \times 10^{-5}\) and for the second stage is \(1 \times 10^{-5}\).

\section{Generative Model Details}
\label{sec:gen}

\paragraph{Architecture Details}
For text-to-video generation, we employ Qwen2.5-VL~\cite{Qwen2VL} as the text encoder, with text conditions injected via cross-attention. Each DiT has 1.3B parameters with a hidden dimension of 1536, 20 blocks, and 16 attention heads.
For class-to-video generation, the DiT architecture consists of a 1024-dimensional hidden state, 24 blocks, and 16 attention heads.

\paragraph{Training Details}

\begin{figure*}
    \centering
    \includegraphics[width=0.99\linewidth]{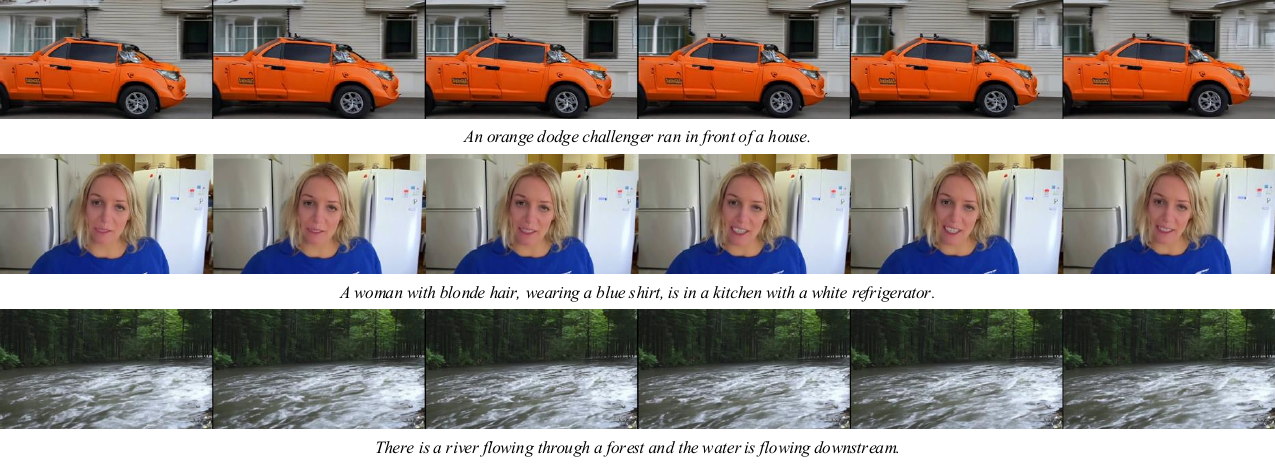}
    \vspace{-1em}
    \caption{Text-to-video results of our latent diffusion model trained on the structural latents of our autoencoder.}
    \label{fig:ldm2d}
\end{figure*}

For text-to-video generation, we employ Qwen2.5-VL~\cite{Qwen2VL} as the text encoder, integrating text features via cross-attention. To ensure training efficiency and effectiveness, we adopt a two-stage strategy. In the first stage, as the size of the structural latent is much smaller than 1D latents, we train DiT  exclusively on structural latents using 48 80G GPUs (per-GPU batch size of 32) for 300K iterations, taking approximately 19 days. As demonstrated in~\cref{fig:ldm2d}, the training of this stage leads to coherent synthesized videos across diverse scenes since the structural latents alone successfully capture sufficient low-frequency semantic information and spatial constraints. In the second stage, the DiT is further trained on both structural and 1D latents with a per-GPU batch size of 8 for 350K iterations, and the corresponding results are shown in~\cref{fig:ldm}.
For class-to-video generation, DiT is directly trained on the full latent space with global batch size of $24 \times 16$ for 800K training iterations.

\section{Autoencoder Adaptation}

\begin{figure}[t]
    \centering
    \begin{subfigure}{0.46\textwidth}
        \centering
        \includegraphics[width=1.0\textwidth]{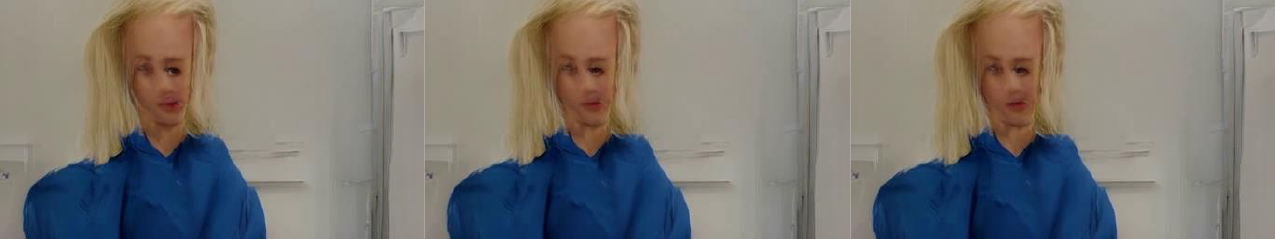}
        \vspace{-1.5em}
        \caption{Pure 1D Latents}
        \label{fig:pure1d}
    \end{subfigure}
    \hfill
    \begin{subfigure}{0.46\textwidth}
        \centering
        \includegraphics[width=1.0\textwidth]{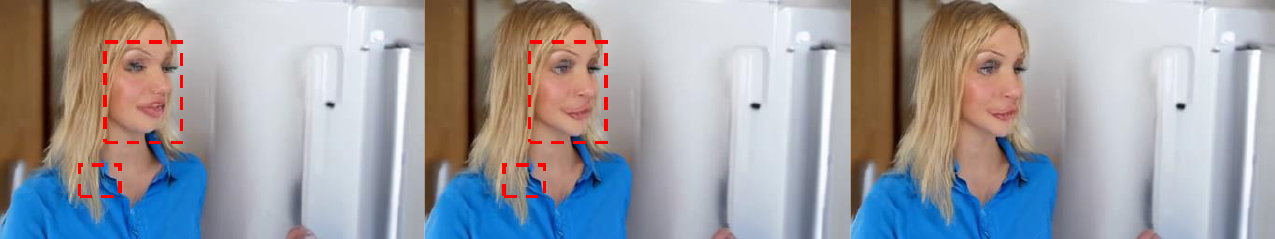}
        \vspace{-1.5em}
        \caption{Hybrid Latents (First-frame Structural)}
        \label{fig:1d2dseparate}
    \end{subfigure}
    \hfill
    \begin{subfigure}{0.46\textwidth}
        \centering
        \includegraphics[width=1.0\textwidth]{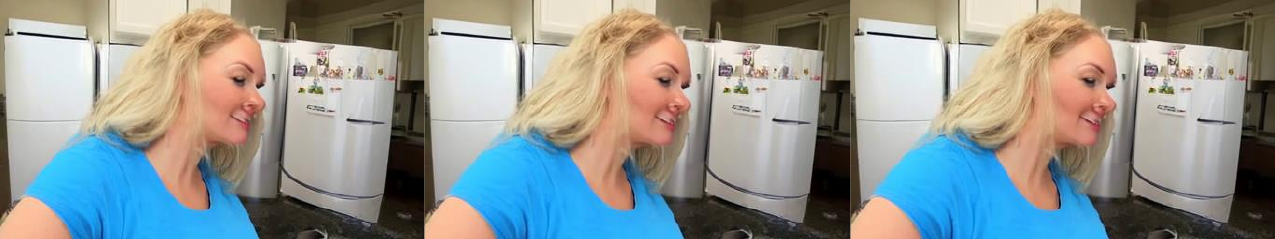}
        \vspace{-1.5em}
        \caption{3D Structural Latents}
        \label{fig:purest}
    \end{subfigure}
    
    \vspace{-.5em}
    \caption{
        \textbf{Three continuous frames generated across different latent spaces.} 
        (a) Results in a pure 1D latent space, exhibiting distorted spatial layouts. 
        (b) Results in a hybrid latent space where only the first frame is structural. The subsequent frames show abrupt transitions and temporal discontinuity (marked by red dashed box). 
        (c) Results in the original 3D structural latent space, maintaining spatiotemporal consistency.
    }
    \label{fig:ablat2vres}
\end{figure}

\begin{figure}[t]
    \centering
    \begin{subfigure}{0.46\textwidth}
        \centering
        \includegraphics[width=1.0\textwidth]{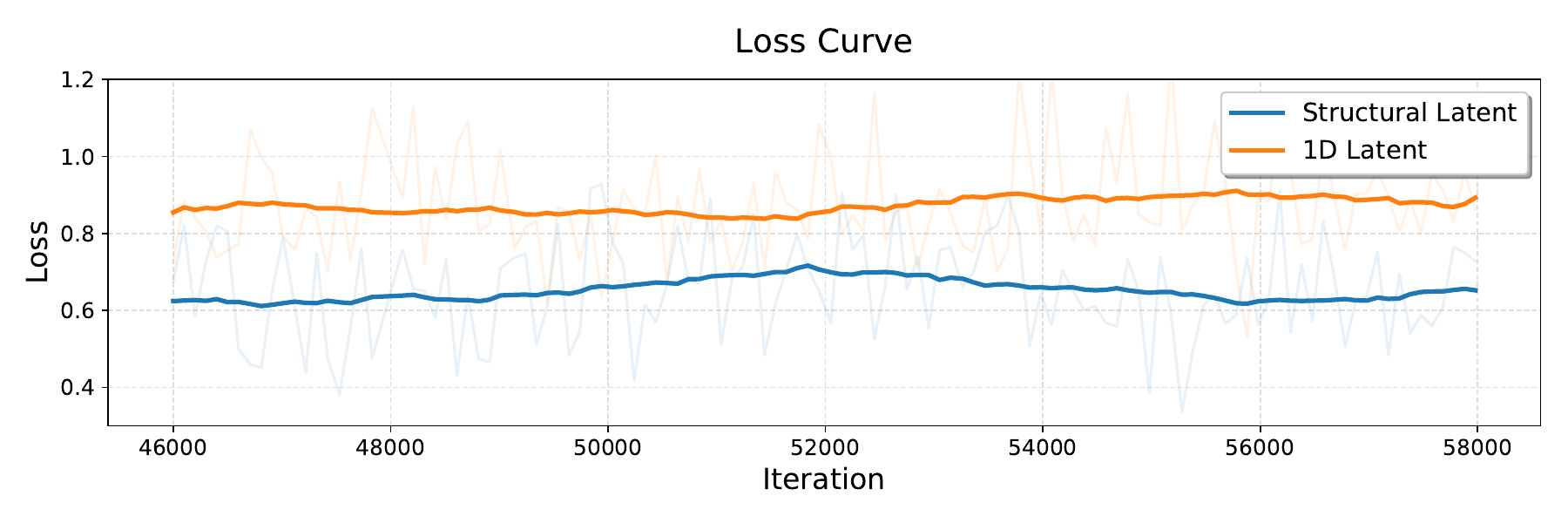}
        \vspace{-1.5em}
        \caption{Before alignment}
        \label{fig:loss_before}
    \end{subfigure}
    \hfill
    \begin{subfigure}{0.46\textwidth}
        \centering
        \includegraphics[width=1.0\textwidth]{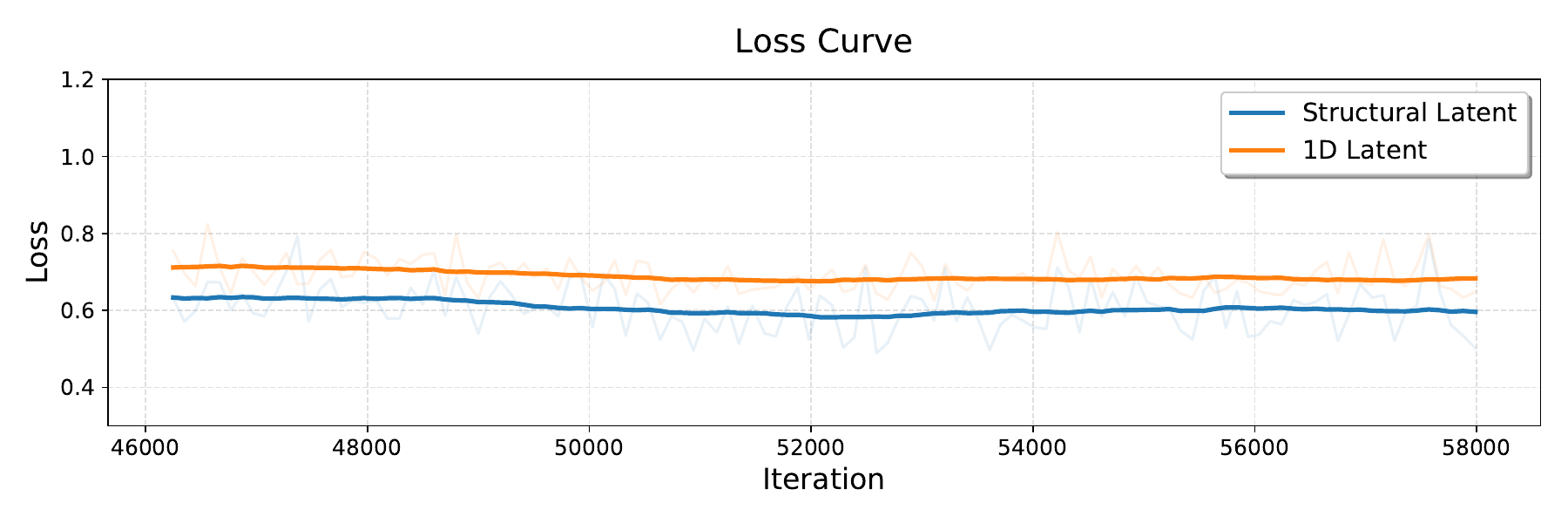}
        \vspace{-1.5em}
        \caption{After alignment}
        \label{fig:loss_after}
    \end{subfigure}
    
    \vspace{-.5em}
    \caption{Effect of latent alignment on training process. (a) Without alignment, the loss curves of different latents exhibit divergence. (b) The proposed alignment mechanism leads to more consistent loss curves.}
    \label{fig:lossalign}
\end{figure}

\paragraph{Discrepancy between Structural and 1D Latents.} As discussed in \cref{sec:gen}, we initially train the video diffusion model exclusively on structural latents to efficiently establish a pretrained video generation model, subsequently incorporating 1D latents. We observe that although both latent types originate from the same Transformer blocks, they exhibit a representation discrepancy. Unlike structural latents, which are derived directly from ViT outputs and possess inherent spatial priors, 1D latents emerge from learnable queries that lack such locality constraints. This discrepancy manifests as various issues, \eg, different statistics, unbalanced loss scales, and distinct visual artifacts. We conduct the following experiments for analysis:

\textit{First}, we fine-tune a variant of One-DVA utilizing only 1D latents (excluding the latent alignment loss). When transferring a diffusion model pre-trained on structural latents to this pure 1D latent space, the model fails to capture coherent spatial structures. Despite an extensive training phase covering over 13 million samples ($105\text{K iterations} \times 128 \text{ global batch size}$), the generated character remains structurally distorted, as shown in \cref{fig:pure1d}. This indicates that 1D latents encode information in a manner different from their structural counterparts. Notably, this failure occurs despite the autoencoder achieving a PSNR of $33.61$, which is sufficient for high-quality reconstruction.
\textit{Furthermore}, we evaluate a hybrid configuration where structural latents encode only the first frame, while 1D latents encode the remainder. We fine-tune the structural-latent-based diffusion model on this configuration using 28 million samples ($290\text{K iterations} \times 96 \text{ global batch size}$). As shown in \cref{fig:1d2dseparate}, while a solid structural foundation is established, subsequent frames suffer from abrupt transitions and temporal discontinuity. 
This suggests that although the structural latent provides a coarse layout for the entire video clip, it is still insufficient to enforce consistency with the generated 1D latent.
\textit{Moreover}, a standard 3D ViT-based autoencoder (utilizing 3D structural latents with a $4\times16\times16$ compression ratio) produces stable results with preserved spatial integrity. This confirms that the structural latent framework is inherently robust, and the observed limitations are specifically tied to the unique behavior of the 1D latents.

Therefore, it is imperative to align the 1D latents with the structural latents. By directly injecting the structural priors into the 1D latent space, we can ensure a consistent spatial correspondence, thereby guaranteeing that each learnable query encodes meaningful and structural information.

\begin{figure*}
    \centering
    \begin{subfigure}[b]{\textwidth}
        \centering
        \includegraphics[width=0.89\textwidth]{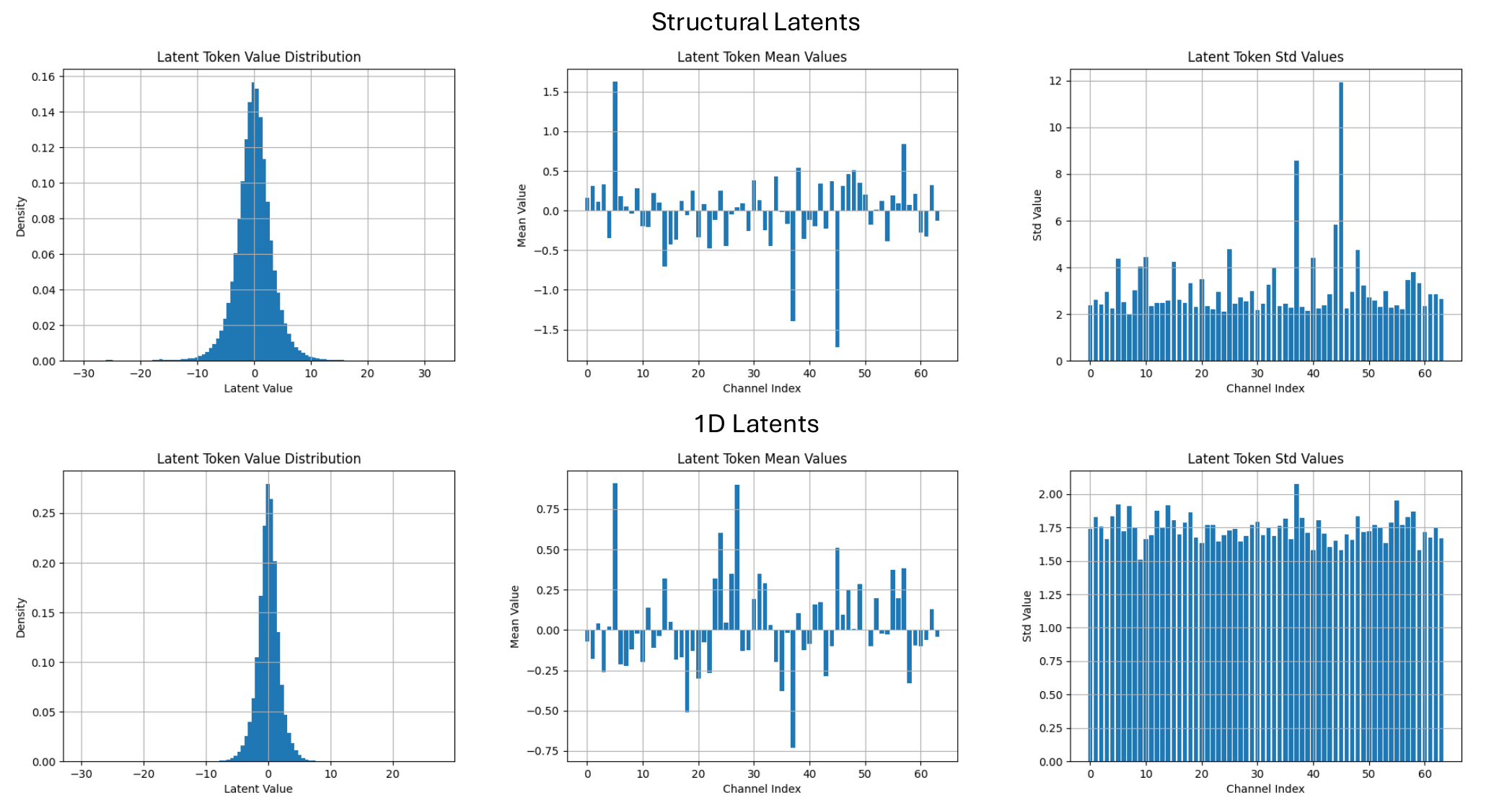}
        \caption{The statistics without the latent alignment process}
        \label{fig:stawoalign}
    \end{subfigure}
    
    \hfill
    
    \begin{subfigure}[b]{\textwidth}
        \centering
        \includegraphics[width=0.89\textwidth]{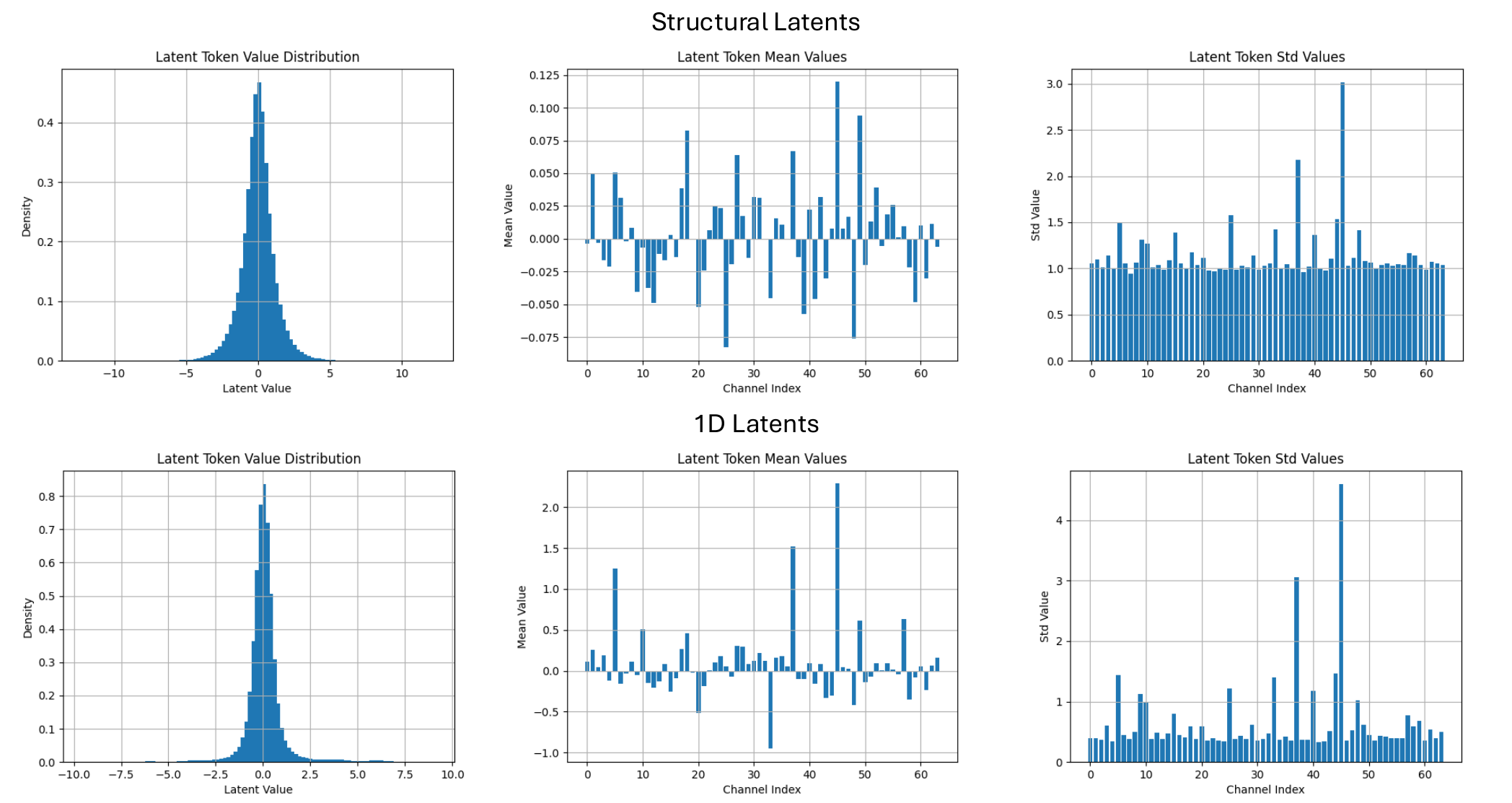}
        \caption{The statistics after the latent alignment process}
        \label{fig:stawalign}
    \end{subfigure}
    
    \caption{The statistics of the latents provided by One-DVA.}
    \label{fig:stalatent}
\end{figure*}

\paragraph{Latent Space Alignment}
As detailed in~\cref{sec:dit}, we regularize the 1D latents via a self-alignment loss to enforce the structural prior. This is achieved by aligning the latents with their best-matching structural counterparts which naturally exhibit smooth and low-frequency characteristics. During this phase, we also increase the KL loss weight for lower latent variance~\cite{bowman2016generating}. As shown in~\cref{tab:align}, a regularization weight of $0.01$ maintains reconstruction fidelity without compromise. 
Such distributional alignment facilitates the learning process for the DiT with channel latent norm, as reflected in the more consistent loss curve shown in~\cref{fig:lossalign}.
Furthermore, we plot the statistics of the latents in~\cref{fig:stawoalign} and \cref{fig:stawalign}. Our analysis reveals that the self-alignment mechanism leads to more consistent statistics across the latent space. For example, the indices of channels exhibiting high variance become nearly identical, indicating improved distributional alignment between the structural and 1D latents.

\begin{table}[t]
    \centering
    \small
    \setlength{\tabcolsep}{3pt}
    \begin{tabular}{lrr|cc}
    \toprule
    Method & Weight & Iters & rFVD ($\downarrow$) & PSNR ($\uparrow$) \\
    \midrule
    One-DVA & / & 797K & 56.96 & 36.48 \\
    \multirow{2}{*}{+ Self-Align Loss} & 0.1 & + 317K & 72.66 & 35.83 \\
                                      & 0.01 & + 135K~ & 59.16 & 36.55 \\
    \bottomrule
    \end{tabular}
    \vspace{-.75em}
    \caption{Reconstruction with self-alignment regularization.}
    \label{tab:align}
\end{table}

\begin{figure}[t]
    \centering
    \begin{subfigure}{0.46\textwidth}
        \centering
        \includegraphics[width=1.0\textwidth]{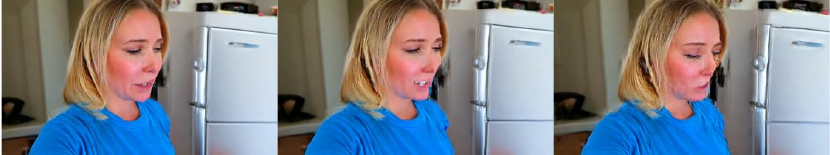}
        \vspace{-1.5em}
        \caption{Decoded frames without decoder finetuning}
        \label{fig:artifacts}
    \end{subfigure}
    \hfill
    \begin{subfigure}{0.46\textwidth}
        \centering
        \includegraphics[width=1.0\textwidth]{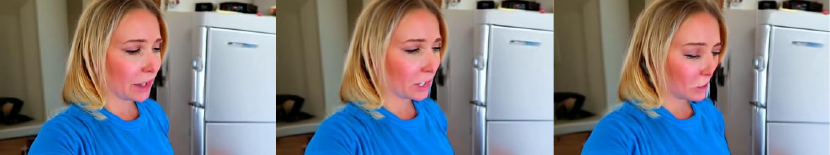}
        \vspace{-1.5em}
        \caption{Decoded frames with decoder finetuning}
        \label{fig:postdec}
    \end{subfigure}
    
    \vspace{-.5em}
    \caption{Visual impact of decoder fine-tuning. (a) Without the finetuning, prediction errors manifest as prominent patch-like artifacts and blocky irregularities on surfaces such as human faces. (b) By post-training the One-DVA decoder on predicted latents, these artifacts are successfully eliminated, significantly enhancing visual smoothness.}
\end{figure}

\paragraph{Decoder Fine-tuning}
We observe that training a diffusion model directly on a combination of structural and 1D latents often results in prominent patch-like artifacts, as shown in~\cref{fig:artifacts}. Even when the optimization of the loss curve appears aligned, 1D latents introduce more obvious artifacts compared to structural latents. To mitigate this, we propose post-training the pixel-space decoder with the well-trained latent diffusion model (LDM) to eliminate these artifacts, an approach supported by the following theoretical intuition. Using the predicted velocity $\bar{\bm{v}}_{\psi}(\bm{z}_t,t,\bm{c})$ for simplicity, the estimated clean latent $\hat{\bm{z}}_0$ can be derived as:
\begin{equation}
\begin{aligned}
    \hat{\bm{z}}_0 &= {\bm{z}}_t -t \cdot \bar{\bm{v}}_{\psi}(\bm{z}_t,t,\bm{c}) \\
    &= (1-t) \cdot {\bm{z}}_0 + t \cdot \bm{\epsilon} -t \cdot \bar{\bm{v}}_{\psi}(\bm{z}_t,t,\bm{c}) \\
    &= {\bm{z}}_0 + t \big[ (\bm{\epsilon} - \bm{z}_0) - \bar{\bm{v}}_{\psi}(\bm{z}_t,t,\bm{c}) \big] ,  \\
    &\text{where}~~ t \sim \mathcal{U}(0, 1), ~~ \bm{\epsilon} \sim \mathcal{N}(\bm{0}, \bm{I}) .
\end{aligned}
\end{equation}
The term $t \big[ (\bm{\epsilon} - \bm{z}_0) - \bar{\bm{v}}_{\psi}(\bm{z}_t,t,\bm{c}) \big]$ can be viewed as a disturbance to the ground-truth latent $\bm{z}_0$, reflecting the gap between the true and predicted velocities. By fine-tuning the decoder using the predicted $\hat{\bm{z}}_0$ as a condition—shifting the mapping from $\mathcal{D}_{\theta}(\bm{x}_s, s, \bm{z}_0)$ to $\mathcal{D}_{\theta}(\bm{x}_s, s, \hat{\bm{z}}_0)$, the model learns to adapt to the training error of the LDM. As illustrated in~\cref{fig:postdec}, we fine-tune the decoder with a batch size of 8 for 40K iterations. These patch-like artifacts are successfully eliminated, leading to enhanced smoothness. Quantitative results in~\cref{tab:c2v} further confirm that this adaptation yields benefits for generation quality.

\section{Further Analysis on Autoencoder}

\begin{figure}[t]
    \centering
    \includegraphics[width=0.98\linewidth]{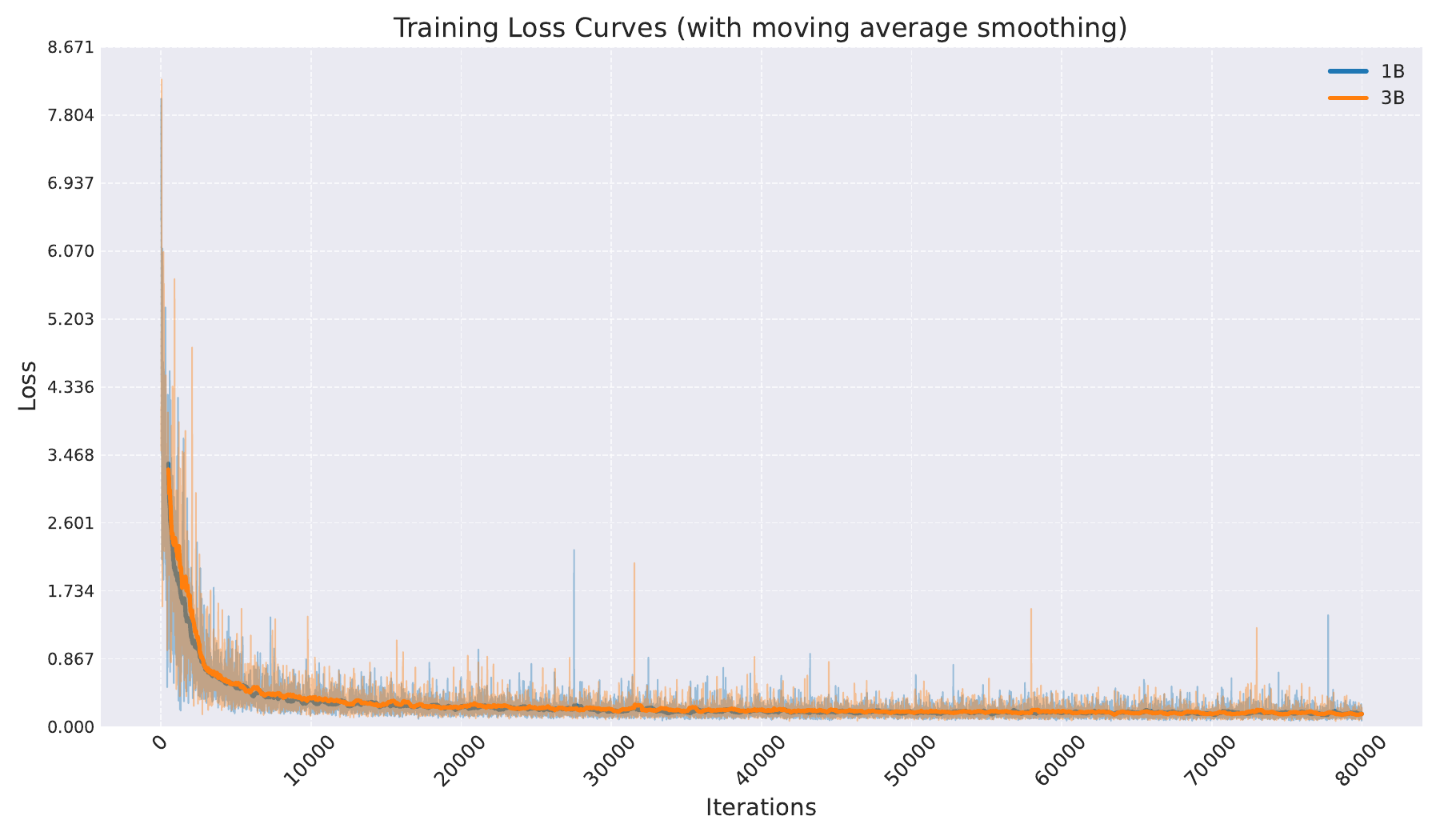}
    \vspace{-0.5em}
    \caption{Loss curves for autoencoders with 1B and 3B parameters. The two curves remain extremely close for the entire training process.}
    \label{fig:ae_scaling}
\end{figure}

\paragraph{Scaling the Autoencoder}
As our autoencoder adopts a transformer-based architecture, we investigate the effect of model scaling. We train variants with 1B and 3B parameters under identical settings and observe that the loss curves are nearly overlapping throughout training, as shown in ~\cref{fig:ae_scaling}. This phenomenon is in line with the finding in~\cite{xiong2025gigatok} where scaling autoencoders beyond 1B parameters brings little improvement in reconstruction. We attribute this phenomenon to the relative simplicity of the reconstruction objective, which appears insufficiently challenging to fully leverage the additional capacity of larger models. Accordingly, we select the 1B-parameter autoencoder as our final model, as a larger variant shows similar reconstruction loss at significantly higher computational cost.

\paragraph{Post-training with GAN Loss} 
We further explore whether introducing a GAN loss~\cite{gan} after the first pretraining stage can improve perceptual quality. As shown in \cref{tab:gan}, even additional 5K iterations of GAN training significantly degrade both rFVD ({67.56} $\to$ {75.48}) and PSNR ({36.02} $\to$ {35.67}). Consequently, we avoid the GAN-based post-training in our framework.

\begin{table}[t]
    \centering
    \setlength{\tabcolsep}{3pt}
    \begin{tabular}{lr|cc}
    \toprule
    Method & Iters & rFVD ($\downarrow$) & PSNR ($\uparrow$) \\
    \midrule
    Pretrained & 415K & 67.56 & 36.02 \\
     + GAN Post-training & + 5K & 75.48 &  35.67 \\
    \bottomrule
    \end{tabular}
    \vspace{-.75em}
    \caption{Study on GAN-based post-training. Applying adversarial training after the pretraining stage harms both rFVD and PSNR.}
    \label{tab:gan}
\end{table}

\section{Limitation and Future Work}
Although One-DVA achieves adaptive compression and high-fidelity reconstruction, several directions remain for further exploration. Currently, while our architecture has the potential to be compatible with streaming generation (\eg, utilizing overlapping spatial-temporal windows, similar to Magi-1~\cite{teng2025magi}, to enable long video modeling), this feature has yet to be fully realized in the experiments. Furthermore, while we employ random sampling to determine token counts during training, identifying the theoretically optimal token length for varying video complexities remains an open question, necessitating a move beyond purely empirical estimations~\cite{li2025learning}. It is also worth noting that the decoder in One-DVA serves not merely to provide supervision to the encoder, but as a critical pixel-space diffusion refiner during inference~\cite{wu2025hunyuanvideo} which can be integrated with features like super-resolution. Moreover, we envision incorporating pre-trained foundation models (\eg, CLIP~\cite{clip}) to develop a more semantically grounded foundation autoencoder with variational or multi-scale encoding capabilities. We also aim to explore an all-in-one pixel-space diffusion decoder that integrates reconstruction and conditional generative tasks (\eg, text/image-to-video) within a single framework. Such an architecture would eliminate the need for a separate latent diffusion model, paving the way toward a truly end-to-end, efficient, and semantically aligned video foundation model.

\end{document}